%% file: main.tex
\documentclass[runningheads]{llncs}

\usepackage{eccv}

\usepackage{eccvabbrv}

\usepackage{graphicx}
\usepackage{booktabs}

\usepackage{multirow}
\usepackage{multicol}
\usepackage{pifont}
\usepackage{color, colortbl}
\usepackage{lipsum}  
\usepackage{tabularx} %

\usepackage[accsupp]{axessibility}  %

\usepackage[pagebackref,breaklinks,colorlinks,citecolor=eccvblue]{hyperref}

\usepackage{orcidlink}

\definecolor{darkergreen}{RGB}{21, 152, 56}
\definecolor{red2}{RGB}{252, 54, 65}
\newcommand{\yesmark}{\textcolor{darkergreen}{\ding{52}}}
\newcommand{\nomark}{\textcolor{red2}{\ding{56}}}
\newcommand{\methodname}{SeiT\texttt{++}\xspace}

\begin{document}

\title{\methodname: Masked Token Modeling Improves Storage-efficient Training}

\titlerunning{\methodname}

\author{
Minhyun Lee$^{1,*,\dagger}$ \quad Song Park$^{2,*}$ \quad Byeongho Heo$^2$ \quad Dongyoon Han$^2$ \quad Hyunjung Shim$^3$ \\
{\small $^*$ Equal contribution} \\
{\small $^\dagger$ Work done during an internship at NAVER AI Lab}
\vspace{0.6em} \\%
{$^1$ Yonsei University \quad $^{2}$ NAVER AI Lab \quad $^{3}$ KAIST AI} \\%
}

\authorrunning{Lee et al.}

\institute{}

\maketitle

\begin{abstract}
Recent advancements in Deep Neural Network (DNN) models have significantly improved performance across computer vision tasks. However, achieving highly generalizable and high-performing vision models requires expansive datasets, resulting in significant storage requirements. This storage challenge is a critical bottleneck for scaling up models. A recent breakthrough by SeiT proposed the use of Vector-Quantized (VQ) feature vectors (\ie, tokens) as network inputs for vision classification. This approach achieved 90\% of the performance of a model trained on full-pixel images with only 1\% of the storage. While SeiT needs labeled data, its potential in scenarios beyond fully supervised learning remains largely untapped. In this paper, we extend SeiT by integrating Masked Token Modeling (MTM) for self-supervised pre-training. Recognizing that self-supervised approaches often demand more data due to the lack of labels, we introduce TokenAdapt and ColorAdapt. These methods facilitate comprehensive token-friendly data augmentation, effectively addressing the increased data requirements of self-supervised learning. We evaluate our approach across various scenarios, including storage-efficient ImageNet-1k classification, fine-grained classification, ADE-20k semantic segmentation, and robustness benchmarks. Experimental results demonstrate consistent performance improvement in diverse experiments, validating the effectiveness of our method. Code is available at \url{https://github.com/naver-ai/seit}.

\end{abstract}

\section{Introduction}
\label{sec:intro}

Recent advancements in Deep Neural Network (DNN) models~\cite{ViT, deit} have significantly improved their performance across various computer vision tasks. However, achieving highly generalizable and high-performing vision models demands an extensive dataset, often containing billions of data points~\cite{jia2021scaling, singh2022revisiting, mahajan2018exploring} with large storage requirements. For example, the LAION-5B~\cite{laion} dataset requires a storage size of 240TB. This storage challenge emerges as a critical bottleneck for scaling up vision models. Consequently, optimizing storage efficiency becomes important for the practical advancement of scaled-up vision models.

\begin{figure}[!t]
    \centering
    \includegraphics[width=0.8\linewidth]{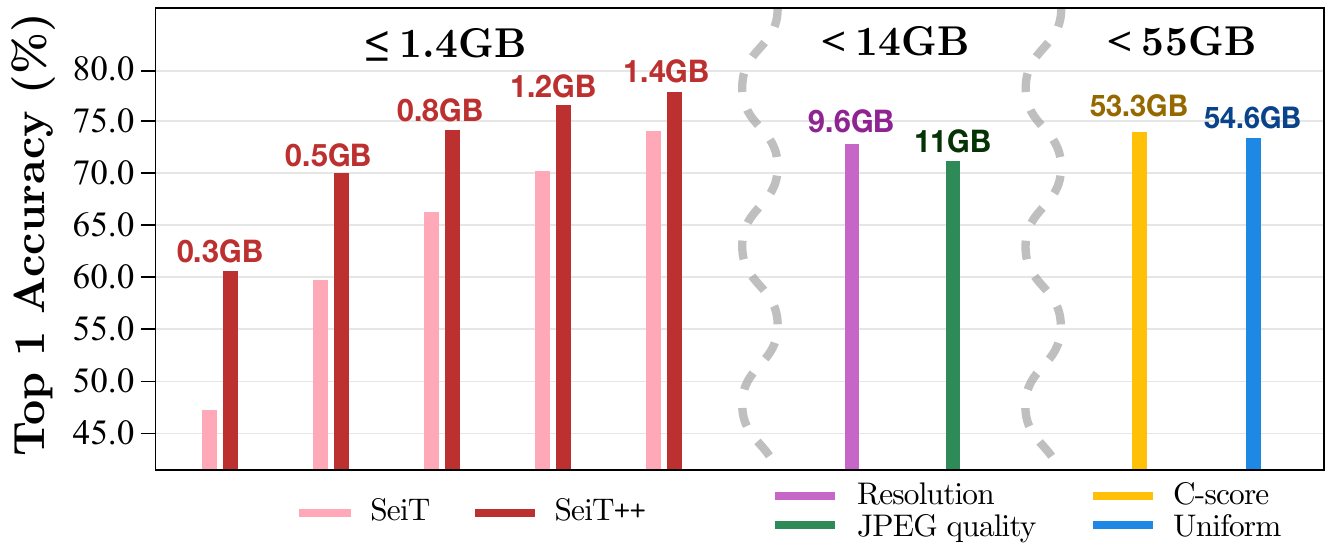}
    \vspace{-.5em}
    \caption{\small \textbf{Over 70\% top-1 accuracy is achievable with just 1GB data.} On ImageNet-1k, we visualize the trade-off of training data storage vs. top-1 accuracy using the fixed ViT-B/16 for a controlled comparison. Each accuracy metric is individually trained with different data types. Consider that while the entire ImageNet-1k dataset requires approximately 140GB for training with images, our approach demonstrates significant storage efficiency over competitors.}
    \label{fig:main}
    \vspace{-1em}
\end{figure}

Images require more storage than texts, due to their intrinsic complexity in converting continuous natural light signals into discrete pixels for computer representation. However, a considerable portion of image data corresponds to redundant or unnecessary details, such as repeated patterns or imperceptible details~\cite{rombach2022latentdiffusion}. These properties lead to large storage requirements and also induce negative impacts on visual recognition tasks as mentioned in~\cite{croce2020reliable,goodfellow2014explaining,madry2017towards,bahng2020learning,geirhos2018imagenet,scimeca2021shortcut}.

Therefore, attempts have been made to reduce storage requirements by optimizing the efficiency of image data. The standard approach to reduce storage requirements is to decrease the total number of training data by removing less important samples~\cite{paul2021deep, coleman2019selection, jiang2020characterizing} or condensing the data into smaller synthetic sets~\cite{zhao2021dataset, zhao2020dataset}. However, as SeiT~\cite{seit} illustrates, this approach can lead to a significant performance drop when compared to using the entire dataset or become impractical for large-scale datasets due to their complexity. Alternatively, some methods aim to reduce the size of each image through compression techniques or resolution adjustments~\cite{laion}. However, they still face a considerable performance gap with the original dataset. Recently, SeiT~\cite{seit} has made a significant breakthrough in this area by enabling the direct learning of ViTs~\cite{ViT} from discrete representations (\ie, tokens) pre-extracted by an offline tokenizer~\cite{ViT-VQGAN}. By using vector-quantized features instead of traditional pixel images, SeiT achieves 90\% of the performance of a full storage model while requiring only 1\% of the storage.

While SeiT~\cite{seit} has demonstrated the potential of token-based training, exploration beyond fully supervised learning scenarios remains an untapped area. To leverage the advantages offered by large-scale data beyond human-annotated datasets, it is imperative to explore label-free learning methods. In this paper, we investigate the compatibility of SeiT with self-supervised learning approaches in token-based frameworks. Among many different self-supervised algorithms, Masked Image Modeling (MIM)~\cite{mae, simmim} derived from Masked Language Modeling~\cite{devlin2018bert} has brought significant success and has quickly become a popular approach for learning visual representations. Inspired by this, we adopt Masked Token Modeling (MTM) as a pre-training method in token-based frameworks and validate its effectiveness for storage efficiency on various benchmarks. To the best of our knowledge, this is the first approach for learning an MLM-based model directly from offline tokens, offering the potential to optimize storage efficiency for large-scale training tasks.

It is worth noting that self-supervised approaches often demand more data due to the lack of labels. MIMs also utilize data augmentation techniques to the core, including a random resized crop and horizontal flip. As previous studies~\cite{randaug,autoaug,deit} have highlighted, data augmentation plays a crucial role in large-scale model training and is a fundamental building block that is widely used in various training recipes. However, applying existing data augmentation techniques designed for pixel images to tokens poses a significant challenge due to the input domain shift. Given its importance in DNN-based models, it is crucial to address the challenges posed by applying data augmentation to tokens. This resolution holds the key to overcoming major performance bottlenecks and improving the storage efficiency of token-based frameworks.

To this end, we introduce TokenAdapt and ColorAdapt, two novel augmentation strategies designed for token-based training. TokenAdapt facilitates the integration of traditional pixel-based augmentation policies by transforming token embeddings into a feature space compatible with pixel-based augmentations. After applying augmentation to the transformed token embeddings, the augmented token embeddings are then reverted to their original token space. ColorAdapt, inspired by Adaptive Instance Normalization (AdaIN~\cite{adain}), alters the color attributes of token embeddings while preserving object structure by adjusting their statistics. By integrating TokenAdapt and ColorAdapt into our token augmentation process, we enhance the effectiveness of token-based training across both fully supervised and unsupervised learning scenarios, demonstrating significant improvements in model performance.

Our \methodname framework combines Masked Token Modeling (MTM) with the proposed token augmentation strategies, TokenAdapt and ColorAdapt. We validate the effectiveness of \methodname on various scenarios: (1) storage-efficient ImageNet-1k classification, (2) transfer learning on fine-grained classification, (3) ADE-20k semantic segmentation, and (4) evaluation on robustness benchmarks. As shown in Figure~\ref{fig:main}, \methodname consistently outperforms comparison methods in storage-efficient ImageNet-1k classification with limited storage sizes, achieving a top-1 accuracy of 77.8\% with only 1.4GB of data. These results confirm that token-based training can take advantage of self-supervised learning to improve storage efficiency. We also demonstrate that our proposed token augmentations significantly contribute to storage efficiency by augmenting data effectively. We reveal our token-based augmentation methods basically yield significant boosts in supervised learning and further amplify improvements in MTM learning. Furthermore, We present how our strategies enhance the generalizability of trained models, as evidenced by our experimental results on fine-grained classification and robustness evaluation. Lastly, the applicability of our method to an alternative input format (\eg, DCT coefficients) demonstrates its extensibility.

\section{Related Work}
\label{sec:related-works}

\subsubsection{Learning with Tokenization.}
The use of discrete tokens for training vision models has recently gained attention, particularly in self-supervised representation learning. BeiT~\cite{Beit} employs the Masked Image Modeling (MIM) framework to recover discrete tokens from masked images and CIM~\cite{CIM} utilizes tokens in Corrupted Image Modeling. MAGE~\cite{MAGE} demonstrates the incorporation of tokens from VQGAN~\cite{VQGAN} for both generative training and representation learning within a unified training framework. However, existing studies mainly focus on self-supervised representation learning and rely on online tokenization, converting images to tokens during each iteration of model training. This approach incurs notable memory consumption and computational costs, limiting the full exploitation of the storage efficiency of discrete tokens.

\subsubsection{Storage-efficient Vision Training.}
Storage-efficient vision training aims to reduce the dataset storage size. To this end, dataset distillation (DD)~\cite{wang2018dataset} creates a condensed dataset by leveraging the knowledge of the original dataset. Recent research~\cite{lee2022dataset, rosasco2021distilled, sangermano2022sample, zhao2021dataset, zhao2020dataset} highlights the effectiveness of dataset distillation in improving training efficiency. However, its high computational complexity poses challenges, particularly when applied to large-scale datasets like ImageNet-1k~\cite{IMNET-1k}. Sampling-based methods~\cite{paul2021deep, coleman2019selection, jiang2020characterizing} address this by selecting a subset of the most representative samples from the full dataset. Paul \etal~\cite{paul2021deep} identify important samples very early in training. Jiang \etal~\cite{jiang2020characterizing} utilize a consistency score based on statistics collected during training. SVP~\cite{coleman2019selection} constructs small proxy models to find important samples. However, the diverse sample selection is not guaranteed, particularly in low data regimes, leading to suboptimal performance compared to dataset distillation (DD) methods~\cite{zhao2020dataset}. To tackle these issues, SeiT~\cite{seit} proposes a storage-efficient vision training framework by leveraging the storage efficiency of tokens. They utilize the ViT-VQGAN~\cite{ViT-VQGAN} tokenizer to extract tokens from images and store them for vision model training. As a result, they outperform other storage-efficient training methods, significantly reducing the dataset storage size.

\subsubsection{Data Augmentation.}
Data augmentation serves as a fundamental building block to improve the model performance across diverse computer vision tasks. Various data augmentation techniques have been employed for improved performance~\cite{cutout,randerasing} or model robustness~\cite{ford2019adversarial,lopes2019improving}. Mixup~\cite{mixup} introduces convex combinations of pairs of images and their labels to create a smoother decision boundary. CutMix~\cite{cutmix} replaces certain regions of an image with a patch from another image, enhancing the model's localization ability. AutoAugment~\cite{autoaug} leverages reinforcement learning to automate optimal augmentation policy discovery. RandAugment~\cite{randaug} introduces a potent policy based on random combinations of different data augmentation methods. DeiT~\cite{deit} emphasizes the crucial role of data augmentation in vision transformers for improved model performance. Previous studies extensively explore the effective data augmentation policies for pixel-based vision model training. However, we observe that pixel-based data augmentations are not well-suited for tokens, posing a critical performance bottleneck. Motivated by this, we tackle the data augmentation challenge, which has not been explored in previous studies for vision model training with tokens.

\section{Method}

\begin{figure}[t]
\centering
    \includegraphics[width=\linewidth]{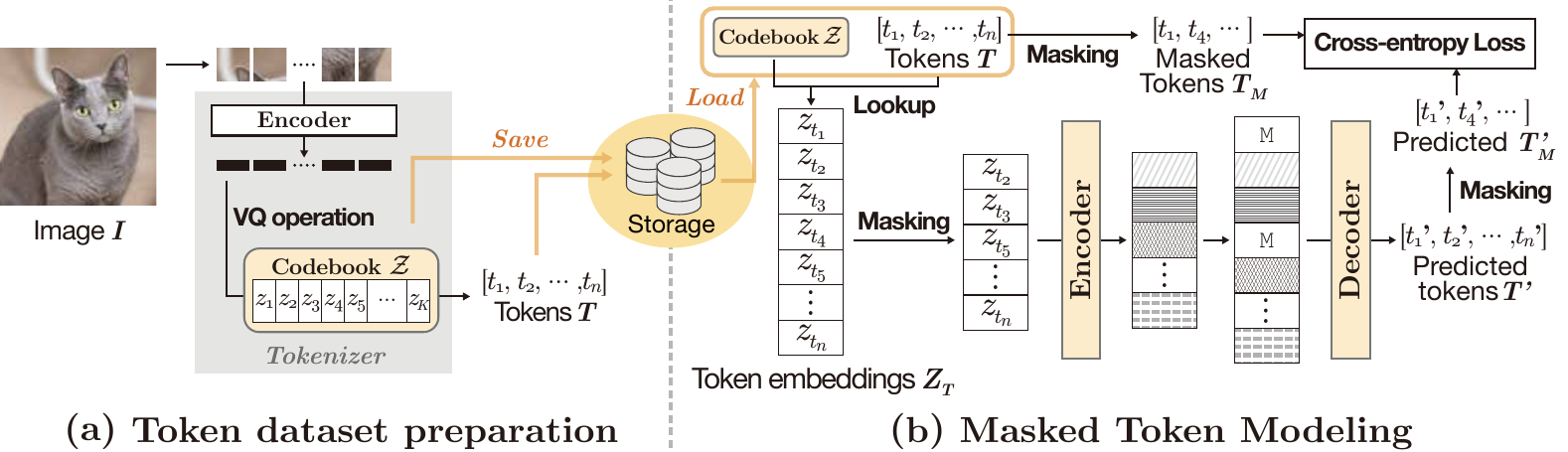}
    \caption{\small \textbf{Masked Token Modeling (MTM) pipeline.} MTM is a self-supervised learning approach in token-based frameworks. (a) The tokenized dataset is saved on storage before model training. Then, (b) using only the pre-stored tokens, a storage-efficient vision model (MTM) is trained without relying on labeled datasets.}
    \label{fig:overview_mtm}
\end{figure}

\subsection{Preliminary: Storage-efficient Vision Training (SeiT)}

\subsubsection{SeiT}\cite{seit} aims to develop a scalable and high-performance vision classifier with minimal storage requirements. It introduces a storage-efficient vision training framework that enables the direct learning of ViT~\cite{ViT} from discrete representations (\ie, tokens) pre-extracted by an offline tokenizer. In addition, SeiT further improves training by incorporating two token-specific augmentations: Token-EDA, drawing inspiration from easy data augmentation for language~\cite{EDA}, and Emb-Noise, which introduces Gaussian noise into discrete representations. By using vector-quantized features instead of pixel images, SeiT retains 90\% of the precision of a model trained on full-pixel images while using only 1\% of the storage space. SeiT consists of two parts: (1) preparing the token dataset and (2) training a model using the token dataset.

\subsubsection{Preparing the Token Dataset.}
SeiT extracts tokens using the ImageNet-1k-trained ViT-VQGAN tokenizer~\cite{ViT-VQGAN} to prepare the token dataset for training. As illustrated in Figure~\ref{fig:overview_mtm}(a), the process begins by transforming each pixel image $I$ into a sequence of tokens $T = [t_1, t_2, ..., t_n]$. Since the tokens are stored in quantized form, each token means an embedding vector in a codebook $\mathcal{Z} = \{z_k \in \mathbb R^d | k = 1, ..., K\}$. \ie, $t_i \in \mathbb{Q}$ where $\mathbb{Q} = \{1, 2, ..., K\}$ for a codebook with size $K$. Here, given the tokens $T$ and the codebook $\mathcal{Z}$, the corresponding token embeddings $Z_T=[z_{t_1}, z_{t_2}, ..., z_{t_n}]$ can be computed by lookup process. This conversion compresses each image $I \in \mathbb{R}^{H \times W \times 3}$ into a more compact token sequence $T \in \mathbb{R}^{n}$, where $n$ is much smaller than $H \times W \times 3$, thus significantly reducing storage requirements. For example, storing the full-pixel ImageNet-1k requires 140GB, but its tokenized form needs only 1.4GB. Notably, the tokens are stored on storage before model training with the codebook $\mathcal{Z}$. Thus, during the training phase, the pre-stored tokens are used directly as input instead of the original pixel images.

\subsubsection{Training Pipeline.}
To train a vision classifier using tokens, SeiT initially loads the token dataset and applies Token-EDA to the tokens. The tokens are then converted to a one-hot format and randomly resized and cropped. Subsequently, the processed one-hot tokens are transformed into token embeddings using a pre-trained codebook. Techniques such as CutMix and Embedding-noise are applied to these token embeddings, and the resulting augmented tensor is fed into the model for training. Through the proposed framework, SeiT demonstrates the potential of token-based training, but it lacks exploration beyond fully supervised learning scenarios. In addition, while token-specific augmentation is proposed, the types of token augmentation are limited and the analysis of data augmentation compared to pixel-based training is insufficient.

\subsection{Masked Token Modeling}
\label{sec:mtm}

In this section, we introduce Masked Token Modeling (MTM), a pre-training method in token-based frameworks without relying on labeled datasets. MTM shares a similar concept to Masked Language Modeling (MLM)~\cite{devlin2018bert}, which works by reconstructing original tokens from their masked versions, employing an autoencoding strategy. Figure~\ref{fig:overview_mtm}(b) presents an overview of how our MTM operates.

\subsubsection{Masking.}
We employ a variable masking ratio that follows a truncated normal distribution, as investigated in previous studies~\cite{muse, maskgit, MAGE}. The tokens are randomly dropped according to the sampled masking ratio, and then the \emph{visible, unmasked} token embeddings are fed into the model. By processing only a subset of token embeddings in the encoder, we significantly reduce the overall pre-training time and memory consumption, which aligns with the findings of MAE~\cite{mae}.

\subsubsection{MTM Encoder and Decoder.}
We use ViT~\cite{ViT} as our encoder because its architecture is capable of processing both the \emph{visible} token embeddings and the original token embeddings, seamlessly accommodating their different lengths. During training, the encoder takes the \emph{visible} token embeddings and outputs latent features that become the decoder input. Before decoding, the latent features are padded to the length of the original token embeddings using mask tokens $\text{M}$. Each mask token is a shared, learned vector indicating the presence of a missing token to be predicted. The decoder then estimates the original tokens $T$ from the padded features.%

 \subsubsection{Training Objective.}
MTM learns to accurately predict the original tokens of each masked token from the visible tokens. In particular, we use a cross-entropy loss between the decoder output and the ground truth tokens as the training objective, which is denoted by $\mathcal{L}_{recon}$:

\begin{equation}
\label{eq:loss_mtm}
\mathcal{L}_{recon} = \text{CE}(T'_{M}, T_{M}),
\end{equation}

\noindent where $T'_{M}$ denotes the predictions of the model for the masked tokens, and $T_{M}$ corresponds to the ground truth tokens for $T'_{M}$. Following MAE~\cite{mae}, this loss is optimized only for the masked tokens.

\begin{figure}[t]
    \centering
    \small
    \includegraphics[width=\linewidth]{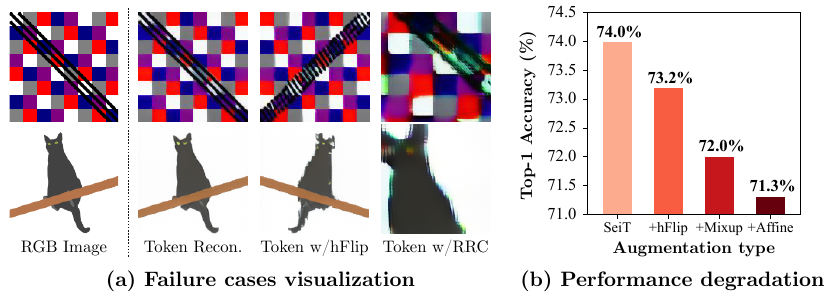}
    \caption{\small \textbf{Data augmentations with tokens.} Each ViT-VQGAN decoded image is reconstructed from a given RGB image after undergoing specific data augmentations. We observe that naively adopting these methods results in incorrect tokenization: 1) Token w/ hFlip demonstrates spatial information collapse during tokenization; 2) Token w/ RRC shows interdependence between neighboring token embeddings. We note the reconstructed images fail to preserve the images' details, suggesting that this incurs ineffectiveness of tokenization. Furthermore, we report top-1 accuracies (ViT-B) on ImageNet-1k according to the augmentations applied during training. All the augmentations are de-facto default training setups for vision transformers~\cite{deit, augreg, swin}.}
    \label{fig:toy}
\vspace{-1em}
\end{figure}

\subsection{Data Augmentation for Tokens}
\label{sec:tokenaug}
While Masked Token Modeling (MTM) improves the storage efficiency of the token-based framework, there is still a performance bottleneck in improving token-based training. Since self-supervised approaches often require more data due to the lack of labels, exploring data augmentation is important to take full advantage of MTM. In this section, we analyze why existing data augmentations designed for pixel images are ineffective in the token domain. We then present TokenAdapt and ColorAdapt as solutions for token-based data augmentation.

\subsubsection{Limited Data Augmentations for Tokens.}
\label{sec:limitation}
We explore the challenges of pixel-based augmentation in the token domain and identify two main reasons: (1) spatial information collapse during tokenization and (2) inter-token dependencies. The discrete tokenization process condenses a $n \times n$ 2D image patch into a single $d$-dimensional 1D vector (\ie, token embedding), causing a collapse of spatial information. This collapse hinders augmentations that rely on spatial details (\eg, horizontal flip). Furthermore, unlike independent image pixels, token embeddings are inter-correlated. This complicates augmentations involving interpolation (\eg, Resize or Mixup~\cite{mixup}), by introducing undesired artifacts.

Figure~\ref{fig:toy} shows the limitations of pixel-based augmentations in the token domain. Applying a horizontal Flip (hFlip) results in the distortion of both the diagonal line (top row) and the cat's silhouette (bottom row) due to spatial information collapse. This occurs because only the inter-token spatial relationship is flipped, while the inner-token spatial arrangement remains unchanged. Augmentations involving interpolation can lead to undesired artifacts, such as color leakage in the case of Random Resized Crop (RRC), due to interdependence between neighboring embeddings. Therefore, we suspect that the direct application of pixel-based augmentations to token embedding may negatively affect token-based training.

\vspace{-1em}
\subsubsection{TokenAdapt.}
The concept of TokenAdapt is to establish a feature space that allows for conventional pixel-based augmentations while facilitating the conversion and inversion of pre-extracted tokens with minimal computational overhead. This approach offers several advantages for our TokenAdapt. Firstly, it enables the direct reuse of well-designed pixel-based data augmentation policies. Additionally, the augmented token can be easily computed through the conversion-augmentation-reverse process. Finally, it ensures that the feature spaces for both the original and augmented tokens are preserved identically. This allows for the seamless use of existing model architectures or pre-trained weights.

\begin{figure}[t]
\centering
    \includegraphics[width=0.85\linewidth]{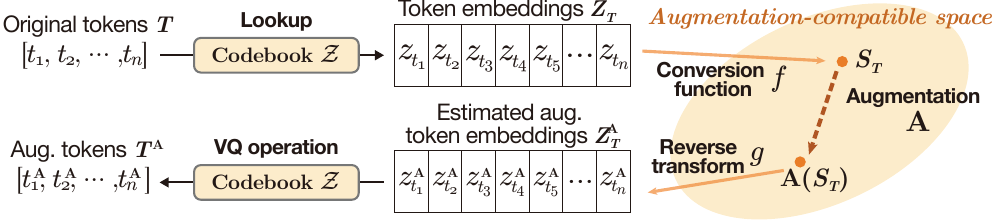}
    \caption{\small \textbf{TokenAdapt processing pipeline.} TokenAdapt aims to enhance the compatibility of token embeddings with pixel-based data augmentations by converting them into augmentation-compatible space, applying augmentations, and reverting them back to the original token embedding space.}
    \label{fig:overview}
    \vspace{-1.5em}
\end{figure}

The goal of TokenAdapt is to design an augmentation function that converts the input tokens $T$ to augmented tokens $T^{\mathbf{A}} = [t^{\mathbf{A}}_1, t^{\mathbf{A}}_2, ..., t^{\mathbf{A}}_n]$. Note that the augmented tokens should also be quantized $t^{\mathbf{A}}_i \in \mathbb{Q}$. Figure~\ref{fig:overview} depicts the overall pipeline of TokenAdapt. 
The initial step of TokenAdapt is transforming token embeddings $Z_T$, computed from $T$ and codebook $\mathcal{Z}$, into a representation $S_T \in \mathbb R^{h \times w \times d}$ in augmentation-compatible space $S$. This transformation is achieved using a conversion function $f$, \ie, $S_T = f(Z_T)$. We apply augmentation $\mathbf{A}$ at augmentation-compatible space $S$ and build an augmented representation $S^\mathbf{A}_T = \mathbf{A}(S_T)$. Then, the augmented representation is reverted to token embedding space by a reverse transformation $g$, $Z^\mathbf{A}_T = g(S^\mathbf{A}_T)$. At last, the augmented tokens in embedding space $Z^\mathbf{A}_T$ are quantized to codebook index as $T^\mathbf{A}=\mathbf{q}_\mathcal{Z}(Z^\mathbf{A}_T)$, where $\mathbf{q}_\mathcal{Z}$ indicates the vector quantization process.
In summary, the augmentation process of TokenAdapt is as follows:

\begin{equation}
\label{augmented_token}
Z^{\mathbf{A}}_T = g(\mathbf{A}(f(Z_T))), \quad T^\mathbf{A} = \mathbf{q}_\mathcal{Z}(Z^{\mathbf{A}}_T).
\end{equation}

\vspace{0.5em}

The conversion function $f$ and the reverse transformation $g$ are learned from paired data of tokens $(T_{x}, T_{\mathbf{A}(x)})$, where $T_{x}$ and $T_{\mathbf{A}(x)}$ denote tokens from an image $x$ and its augmented image $\mathbf{A}(x)$, respectively. We use the cross-entropy loss between $Z_{T_{x}}^{\mathbf{A}}$ and $Z_{T_{\mathbf{A}(x)}}$ as our objective function $\mathcal{L}_{\text{TA}}$ for learning our TokenAdapt module:

\begin{equation}
    \centering
    \mathcal{L}_{\text{TA}} = \text{CE}(Z_{T_{x}}^{\mathbf{A}}, Z_{T_{\mathbf{A}(x)}}).
    \label{objective_function}
\end{equation}

\vspace{0.5em}

\noindent TokenAdapt deals with geometric data augmentations (\eg, flip, resize, crop, affine) and Mixup~\cite{mixup}. These augmentations are commonly used in various training methods, but they can cause a performance drop in token-based training, as shown in Figure~\ref{fig:toy}(b).

Once trained, the TokenAdapt module is used to augment a sequence of tokens $T$ into an augmented sequence $T^{\mathbf{A}}$ across different datasets and downstream tasks. Importantly, the augmentation-compatible feature space, which does not rely on specific semantic knowledge or alignment with downstream tasks, makes it dataset and task-agnostic. The inherent flexibility of TokenAdapt makes it highly generalizable, allowing direct application to other datasets without the need for additional training procedures.

\vspace{-1em}
\subsubsection{ColorAdapt.}
\label{sec:coloradapt}
As witnessed in Cubuk \etal~\cite{autoaug}, color-based augmentations are often more effective than geometric ones in certain datasets and are an important component of data augmentation strategies. For example, color-based augmentations are useful for dealing with various environmental changes caused by different lighting conditions (\eg, day and night) or weather conditions (\eg, snow). However, integrating them into TokenAdapt is challenging because most color-based augmentations are designed to work only in the pixel domain. To address this, we propose ColorAdapt, a color-based token augmentation inspired by the method~\cite{adain}. Formally, the ColorAdapt function $\mathcal{C}$ is defined as:

\begin{equation}
    \centering
    \small
    \mathcal{C}(Z_{T_{1}},Z_{T_{2}}) = \sigma(Z_{T_{2}}) \left(\frac{Z_{T_{1}} - \mu(Z_{T_{1}})}{\sigma(Z_{T_{1}})}\right) + \mu(Z_{T_{2}}),
    \label{eq:token_adain}
\end{equation}

\vspace{0.5em}

\noindent where $\mu(Z_{T})$ and $\sigma(Z_{T})$ are the mean and standard deviation of each token embedding $Z_{T}$ across its spatial dimension, computed for each channel.

\begin{figure}[t]
    \centering
    \small
    \vspace{-1em}
    \includegraphics[width=\linewidth]{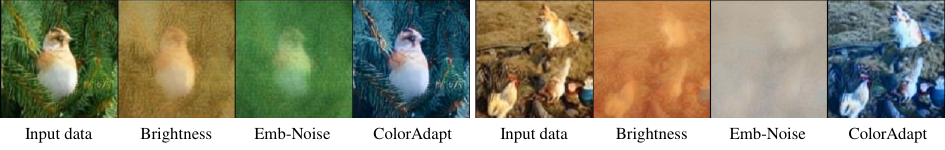}
    \vspace{-2em}
    \caption{\small \textbf{ColorAdapt provides more reasonable color changes.} We present ViT-VQGAN decoded images to verify the quality of tokenizations after color changes. We use the brightness function with a factor of 0.2 following the implementation~\cite{kornia}. Emb-Noise~\cite{seit} is the color-based token augmentation. Notably, our ColorAdapt effectively preserves object structure in contrast to the failure of the counterparts.}
    \label{fig:colorAdapt}
    \vspace{-1em}
\end{figure}

ColorAdapt mimics color-changing augmentations in the pixel domain while preserving the encoded object structure by transforming the global statistics within each channel of the token embedding. Figure~\ref{fig:colorAdapt} shows the images decoded from various color-augmented tokens. This illustrates how ColorAdapt affects the visual properties of the augmented tokens. By incorporating ColorAdapt into our token augmentation strategy, we improve the data diversity, exploring new possibilities for training more robust and generalizable vision models.

\section{Experiment}

In this section, we validate the effectiveness of \methodname on various scenarios. First, we evaluate the performance of our method in the context of storage-efficient ImageNet-1k classification and transfer learning on fine-grained datasets. Then, we demonstrate the adaptability of token-based vision model training to ADE-20k semantic segmentation, showing additional improvements facilitated by \methodname. In addition, we present experimental results illustrating how our strategies contribute to enhanced model generalizability in robustness evaluation. We also explore the applicability of our augmentation strategies to alternative input formats, demonstrating its extensibility. Unless otherwise stated, \methodname refers to the combination of MTM and our token augmentation strategies.

\input{table/main_table}

\subsection{Storage-efficient ImageNet-1k}

Table~\ref{tab:main_results} summarizes the performance and compression ratio of storage-efficient ImageNet-1k classification. Our method is compared with image-based approaches, including uniform random sampling, C-score~\cite{jiang2020characterizing}-based sampling, adjusting image resolution, and adjusting JPEG quality factor. Despite a 1\% compression ratio, when a ViT-B/16 model is pre-trained on tokenized ImageNet-1k using MTM and the ViT model is fine-tuned, our method achieves a 77.8\% top-1 accuracy, outperforming various image-based approaches. In Table~\ref{tab:token-cls}, our method consistently outperforms the token-based learning baseline SeiT by 1.5--6.7\%p in top-1 accuracies. Notably, the effectiveness of our method becomes more pronounced as the available storage size decreases, highlighting its efficiency in low-storage scenarios. Furthermore, \methodname shows more significant performance gains with MTM. For a storage size of 1.4GB, \methodname achieves an improvement of 1.5\%p over SeiT without MTM, while with MTM the improvement increases to 2.7\%p. These results confirm that \methodname effectively takes advantage of MTM.

\input{table/token_cls}
\input{table/fg_cls}

\subsection{Fine-grained Classification}

To further investigate the generalizability of \methodname, we conduct experiments on four different fine-grained datasets: Flowers~\cite{flowers}, StandfordCars~\cite{cars}, iNat-18~\cite{iNat18}, and iNat-19~\cite{iNat19}. Specifically, we tokenize each dataset and then fine-tune the token-trained model (ViT-B) on ImageNet-1k with the tokenized dataset. Since fine-grained datasets require the classification of objects with intricate characteristics, they are particularly sensitive to undesirable noise (\ie, RRC in SeiT) or excessive perturbation (\ie, Emb-Noise) resulting from data augmentation in the token embedding space. Table~\ref{tab:otherdatasets} shows the quantitative performance (top-1 accuracies) on fine-grained datasets. \methodname consistently outperforms the performance of SeiT across various datasets. These results confirm the effectiveness of our proposed method in fine-grained classification.

\input{table/token_seg}

\subsection{ADE-20k Semantic Segmentation}
A limitation of token-based learning is often substantiated by learning dense prediction tasks. We conjecture that our method could facilitate training for those tasks, such as semantic segmentation. We extract tokens from ADE-20k~\cite{ADE20K} and use the pre-trained ViT-B/16 model on tokenized ImageNet-1k to evaluate the effectiveness of our method in semantic segmentation. We then fine-tune the pre-trained model using the tokenized ADE-20k dataset. Table~\ref{tab:seg} reports the mean intersection-over-union (mIoU) as an evaluation metric. The results show a significant performance improvement (+4.2\%p mIoU) over SeiT. These results indicate that \methodname also significantly improves the model performance in the token-based dense prediction task.

\input{table/robustness}

\subsection{Evaluation on Robustness Datasets}
To validate the effectiveness of our augmentation strategies in terms of robustness, we further evaluate the robustness of our method on a set of benchmarks such as adding Gaussian noise~\cite{imagenet-c} or Gaussian blur~\cite{imagenet-c}, ImageNet-R~\cite{imagenet-r}, Sketch~\cite{sketch}, and ObjectNet~\cite{objectnet}. Notably, we adopt \methodname w/o MTM to validate the impact of our augmentation strategies on model robustness. The results in Table~\ref{tab:robustness} show that \methodname consistently outperforms SeiT in terms of robustness accuracy. Previous studies~\cite{chun2020empirical, taori2020measuring} observe that the key to the input pixel robustness depends on the pixel-level augmentations. Since our method enables pixel-based data augmentation to tokens, it achieves higher robustness accuracies than the baseline, which does not effectively exploit the pixel-based data augmentation.

\subsection{Classification with Limited Data}

Data augmentation becomes more critical in limited data scenarios. To assess the impact of our approach in such situations, we conduct experiments on the ImageNet-100 benchmark by randomly sampling data based on varying data ratios. The results (top-1 accuracies), as shown in Table~\ref{tab:limited_data}, consistently demonstrate the superiority of \methodname over SeiT. Notably, the performance gap increases as the amount of training data decreases. With only 13k images, our method shows an improvement of approximately 21\% over the baseline, indicating the effectiveness of our approach, especially in data-scarce scenarios.

\input{table/sub-tables}

\subsection{Ablation Study}
\label{sec:ablation}

\subsubsection{Impact of Each Augmentation Strategy.}
In Table~\ref{tab:abl_component}, we report the impact of proposed token augmentations. In particular, incorporating ColorAdapt yiqualds a 1.0\%p performance improvement, highlighting the positive effect of preserving object structure during model training. Additionally, incorporating TokenAdapt yields an extra 3.1\%p improvement. These results confirm that exploring token augmentations is a key factor in fully exploiting pre-extracted tokens.

\input{table/other_input}

\subsubsection{Extensibility of Our Augmentation Strategy.} 
In this section, we conjecture the broad applicability of our augmentation strategies across various scenarios. To explore its extensibility, we delve into two distinct situations. Firstly, we examine the integration of our method with an alternative tokenizer—the OpenImages-trained VQGAN tokenizer~\cite{VQGAN}, alongside the ImageNet-1k-trained ViT-VQGAN tokenizer~\cite{ViT-VQGAN}. The results in Table~\ref{tab:dct} consistently demonstrate improved performance on ImageNet-100 across different tokenizers. Additionally, we investigate the adaptability of our approach to a different input format by focusing on Discrete Cosine Transform (DCT) coefficients. Inspired by JPEG compression, we apply DCT to each $8 \times 8$ image patch, yielding quantized DCT coefficients utilzed as input data during model training. As shown in Table~\ref{tab:dct}, our approach enhances top-1 accuracy, similar to the tokenizer scenario. This consistent improvement across diverse input formats underscores the universal applicability of our method, highlighting its potential as a widely applicable solution for diverse input representations.

\subsection{Implementation Details}

We use the pre-trained ViT-VQGAN tokenizer~\cite{ViT-VQGAN} for token extraction. For our TokenAdapt module, we employ a single transformer block for conversion and reverse function. The TokenAdapt module is trained for one epoch on ImageNet-1k with a batch size of 128, a learning rate of 0.001 with cosine scheduling, and a weight decay of 0.05. 
We follow the training recipes from MAGE~\cite{MAGE}, SeiT~\cite{seit}, and mmsegmentation~\cite{mmseg} for masked token modeling (MTM), token-based classification, and semantic segmentation, respectively.
Detailed hyperparameters and settings are available in the Appendix.

\section{Conclusion}

In this paper, we investigate the compatibility of SeiT with self-supervised learning approaches in token-based frameworks. Among various self-supervised algorithms, Masked Image Modeling (MIM)~\cite{mae, simmim} has recently gained popularity for learning visual representations. Inspired by this, we extend SeiT by integrating Masked Token Modeling (MTM) for self-supervised pre-training. Additionally, recognizing that self-supervised approaches often require more data due to the lack of labels, we propose simple yet effective token-based augmentation strategies: TokenAdapt and ColorAdapt. TokenAdapt aligns the token embedding space with another embedding space compatible with pixel-based data augmentations. In addition, we introduce ColorAdapt, a color-changing augmentation for tokens. Our experimental results across various scenarios, including storage-efficient ImageNet-1k classification, fine-grained classification, ADE-20k semantic segmentation, and robustness benchmarks, consistently validate the effectiveness of our proposed method. Furthermore, the consistent performance improvement across diverse input formats highlights the potential of our approach as a widely applicable solution for various input representations. \\

\noindent{\textbf{Limitations.}} While \methodname is capable of handling large-scale data without labels with optimized storage requirements, experiments on such large datasets (\eg, ImageNet-21k~\cite{IMNET-1k}) are not contained in this paper due to resource constraints. Nevertheless, the advantages of MTM as self-supervised learning and the integration of our data augmentation strategies can be seamlessly integrated into larger datasets.

\bibliographystyle{splncs04}
\bibliography{main}

\clearpage

\section*{Supplementary Material}

\appendix

\setcounter{section}{0}
\setcounter{table}{0}
\setcounter{figure}{0}
\renewcommand\thefigure{\Alph{figure}}    
\setcounter{figure}{0}  
\renewcommand\thetable{\Alph{table}}    
\setcounter{table}{0}

In this supplementary material, we describe the details of \methodname. Section~\ref{appendix:eff_tokenadapt} examines the linear separability (Section~\ref{appendix:lin_sep}), loss curve analysis (Section~\ref{appendix:loss_curve}), data efficiency (Section~\ref{appendix:data_eff_tokenadapt}), and computational cost (Section~\ref{appendix:cost_tokenadapt}) of the proposed token augmentation strategy. Additionally, Section~\ref{appendix:qual} presents more qualitative results showing the effectiveness of our proposed TokenAdapt and ColorAdapt for augmenting tokens. Lastly, Section~\ref{appendix:details} provides implementation details for our experiments.

\begin{figure}[]
    \centering
    \includegraphics[width=0.8\linewidth]{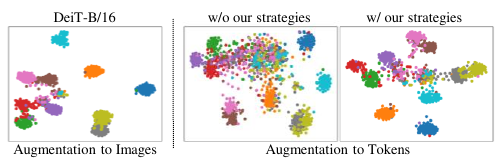}
    \caption{\small \textbf{Linear separability visualization.} We visualize augmented data points in the penultimate layer on ImageNet-1k, randomly sampling 10 classes. Different colors represent distinct classes. Notably, our proposed augmentation strategies exhibit enhanced linear separability compared to the direct application of pixel-based data augmentation to token embedding. This behavior aligns with the observed trend when applying data augmentation to images.}
    \label{fig:tsne}
    \vspace{-2.0em}
\end{figure}

\section{Analysis of TokenAdapt}
\label{appendix:eff_tokenadapt}

\subsection{Linear Separability of TokenAdapt}
\label{appendix:lin_sep}

To demonstrate the efficacy of our token augmentation strategies, we visualize the embedded feature representation in Figure~\ref{fig:tsne}. This illustration offers insight into the linear separability of augmented data points, particularly when pixel-based data augmentation is applied through our proposed method. When pixel-based data augmentation is directly applied to token embedding without our method, there is significant overlap among augmented data points from different classes, indicating low linear separability. This overlap suggests that augmenting tokens without our method leads to substantial shifts in the distribution of training data. Maintaining distributional similarity between clean and augmented data is crucial for model performance, as witnessed in Cubuk \etal~\cite{cubuk2021tradeoffs}, and significant distribution shifts can result in performance degradation. In contrast, our method shows high linear separability, demonstrating its effectiveness in diversifying the data while minimizing the distributional gap between clean and augmented data. Moreover, this behavior aligns with the observed trend in the case of applying data augmentation to images (DeiT-B~\cite{deit} in Figure~\ref{fig:tsne}).

\begin{figure}[t]
    \centering
    \includegraphics[width=\linewidth]{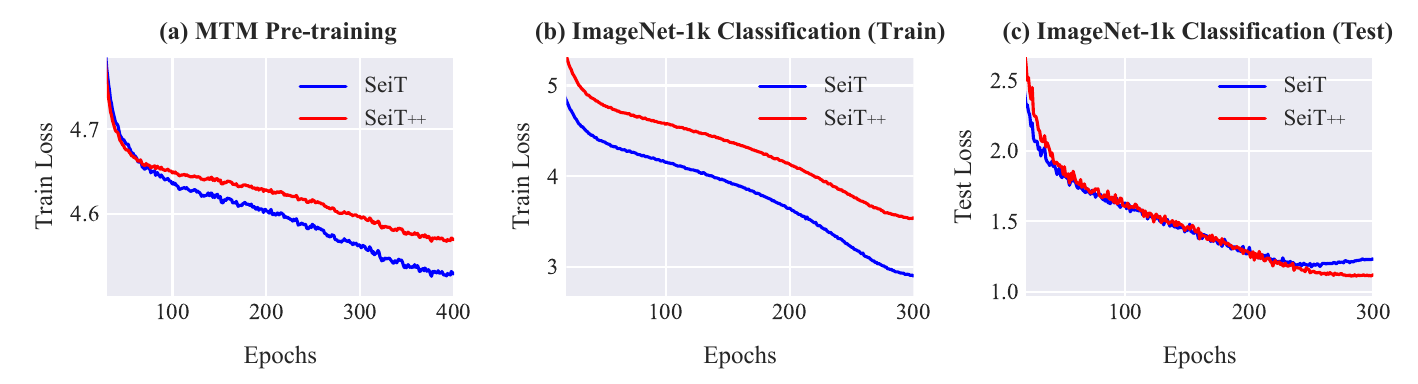}
    \vspace{-1em}
    \caption{\small \textbf{Loss curve visualizations.} We provide loss curves for ViT-B during MTM pre-training and during token-based ImageNet-1k classification for both \methodname and SeiT~\cite{seit}. Notably, \methodname shows more regularized results than SeiT, effectively leveraging pixel-based data augmentations on tokens to avoid overfitting. This evidences that token-based learning requires much stronger data augmentation for a more generalizable representation learning.}
    \label{fig:loss_curve}
\end{figure}

\subsection{Loss Analysis of TokenAdapt}
\label{appendix:loss_curve}

We analyze the influence of our approach on the training of ViT-B for both MTM pre-training and token-based ImageNet-1k classification, as illustrated by the loss curves in Figure~\ref{fig:loss_curve}. Observing the training loss, our analysis shows that \methodname, enhanced by our token augmentation strategy, maintains higher training loss values than SeiT~\cite{seit} in both MTM pre-training and token-based ImageNet-1k classification. This pattern indicates that our method's token augmentation enriches the data diversity, which helps to mitigate overfitting. Meanwhile, Figure~\ref{fig:loss_curve}(c) shows that the test loss aligns with the trends of SeiT. Notably, while SeiT's test loss increases in later training epochs, our approach consistently decreases, highlighting its effectiveness in preventing the overfitting problem.

\begin{table}[t]
\small
\centering
\caption{\small \textbf{Data efficiency of the TokenAdapt module.} We investigate the data efficiency of our TokenAdapt module by training it with varying amounts of training data. We validate the trained TokenAdapt module's efficacy in storage-efficient ImageNet-1k classification. Notably, the TokenAdapt module can be trained even with 100 images, indicating its robustness to data scarcity.}
\begin{tabular}{c@{\hskip 12pt}c@{\hskip 12pt}c@{\hskip 12pt}c@{\hskip 12pt}c@{\hskip 12pt}c}
\toprule
\# of training data & \multirow{2}{*}{100}  & \multirow{2}{*}{1k}   & \multirow{2}{*}{5k}   & \multirow{2}{*}{256k} & \multirow{2}{*}{1281k} \\
(TokenAdapt) &                       &                       &                       &                       &                         \\ \midrule
Top 1 Acc.     & \multirow{2}{*}{75.1} & \multirow{2}{*}{75.4} & \multirow{2}{*}{75.4} & \multirow{2}{*}{75.4} & \multirow{2}{*}{75.5}   \\
(IN-1k CLS.)   &                       &                       &                       &                       &                         \\ \bottomrule
\end{tabular}
\label{tab:abl_tam}
\vspace{-1em}
\end{table}

\subsection{Data Efficiency of TokenAdapt}
\label{appendix:data_eff_tokenadapt}
To investigate the data efficiency of the TokenAdapt module, we explore the impact of reduced training data for the TokenAdapt module training. Specifically, we vary the number of training data for the TokenAdapt module training and subsequently train a token-based ImageNet-1k classification model using the trained TokenAdapt module. Table~\ref{tab:abl_tam} shows that even with a reduced number of training data for our TokenAdapt module, the performance degradation is minimal. Notably, the top-1 accuracy remains more than 1\%p higher than the baseline SeiT (74.0\% top-1 accuracy), even with only 100 training data. This robust performance suggests that our TokenAdapt module is not significantly affected by variations in the training data size.

\begin{table}[t]
\small
\centering
\caption{\small \textbf{Computational costs comparison.} We report computational costs and top-1 accuracy by training ViT-S on ImageNet-100. Baseline$\dagger$ denotes that (1) tokens are decoded to pixel-level images, (2) pixel-based data augmentations (\eg, hFlip, affine transformations, Mixup, and CutMix) are applied to the decoded images, and then (3) augmented images are encoded to tokens back. GPU hours refer to the number of hours required for model training when using V100 4 GPUs.}
\begin{tabular}{l@{\hskip 12pt}l@{\hskip 12pt}l@{\hskip 12pt}l}
\toprule
\multirow{1}{*}{Method}         & \multirow{1}{*}{\#params} & \multicolumn{1}{l}{GPU hours}              & \multirow{1}{*}{Top 1 Acc.} \\ \midrule
SeiT~\cite{seit}                        & 22M                       & \hspace{.2em} 5.9                        & 77.3           \\
SeiT~\cite{seit}$^\dagger$               & 22M+172M                  & 49.2 \textcolor{red}{(+734\%)}                      & 80.1 \textcolor{darkergreen}{(+2.8)}          \\ 
\midrule
Ours                            & 22M+3.7M                  & \hspace{.2em} 6.4 \textcolor{red}{(+8\%)}                        & \textbf{81.4 \textcolor{darkergreen}{(+4.1)}}            \\ \bottomrule
\end{tabular}
\label{tab:cost}
\vspace{-1em}
\end{table}

\subsection{Computational Costs} 
\label{appendix:cost_tokenadapt}

To validate the computational efficiency of TokenAdapt, we compare ViT-S models trained on the tokenized ImageNet-100 with different training strategies (Table~\ref{tab:cost}). Three scenarios are considered: (1) token-based augmentation only (\ie, SeiT), (2) pixel-based augmentations in decoded images (SeiT$^\dagger$), and (3) pixel-based augmentations using our TokenAdapt module. For SeiT$^\dagger$, using the ViT-VQGAN decoder (86M) and encoder (86M) during each forward computation significantly increases training time by over 8 times compared to the original SeiT training, making it impractical for addressing the data augmentation challenge. In contrast, our TokenAdapt module, requiring only 3.7M parameters, increases training time by only 8\% compared to SeiT while achieving a remarkable performance improvement over 3\%p in top-1 accuracy. Notably, TokenAdapt even outperforms SeiT$^\dagger$, indicating that the full decoding-augmentation-encoding process may introduce undesirable noise during tokenization. These results demonstrate the efficient handling of the data augmentation challenge by our TokenAdapt module. Notably, the TokenAdapt module can be removed during the inference.

\section{More Qualitative Examples}
\label{appendix:qual}

\subsubsection{TokenAdapt.}
\begin{figure*}[]
    \centering
    \small
    \includegraphics[width=\linewidth]{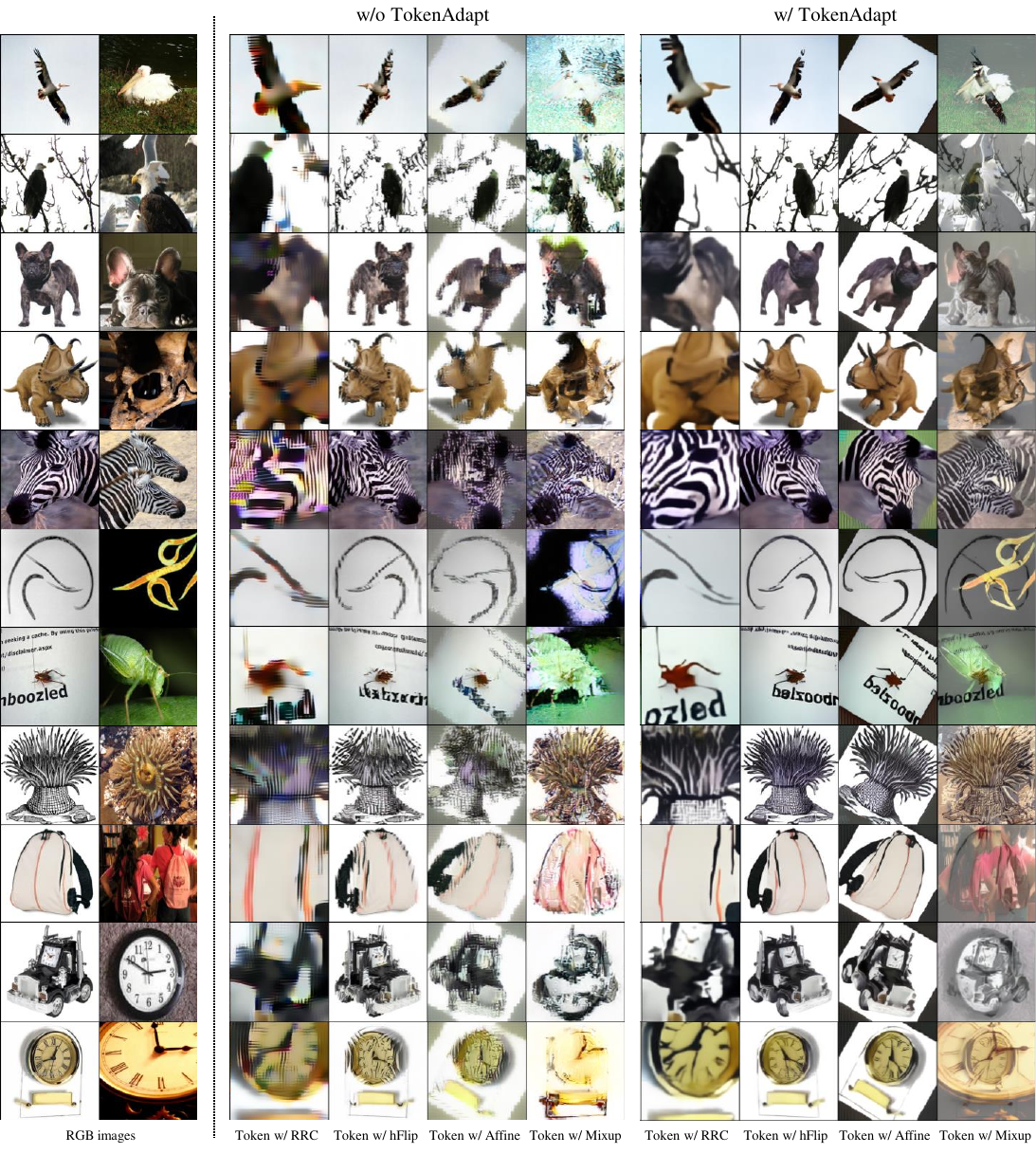}
    \caption{\small \textbf{TokenAdapt provides more reasonable results when augmenting tokens.} We present ViT-VQGAN decoded images to verify the quality of tokenizations after applying pixel-based data augmentations to tokens. The direct application of pixel-based data augmentation to token embedding (w/o TokenAdapt) results in undesired artifacts. In contrast, our TokenAdapt yields more reasonable results. Token w/ Affine indicates the application of affine transformations (\eg, rotation, translation, shear, etc.) to token embedding. For Token w/ Mixup, we mixed the two tokens with a 1:1 ratio (\ie, interpolation ratio $\lambda = 0.5$).}
    \label{supp:tokenAdapt}
\end{figure*}

We demonstrate the effectiveness of the proposed TokenAdapt and ColorAdapt by decoding the augmented tokens into images using the ViT-VQGAN decoder~\cite{ViT-VQGAN}. In Figure~\ref{supp:tokenAdapt}, we compare the direct application of pixel-based data augmentations to token embedding with augmentations using our TokenAdapt module. As mentioned in Section~\textcolor{red}{3.3}, direct application of hFlip or affine transformations leads to disruptions in both the diagonal line and the silhouette of the object due to spatial information collapse. In addition, data augmentations related to interpolation (\eg, Resize or Mixup) can result in undesired artifacts. Mixup, in particular, occasionally causes significant disruption of objects, leading to substantial performance degradation as shown in Figure~\textcolor{red}{3}(b).

The occurrence of these unexpected artifacts makes it difficult to use the data augmentations that are widely used in existing pixel image domains. Moreover, in tasks such as MTM and semantic segmentation, where pixel-level details are essential, such artifacts significantly affect learning stability and model performance. In contrast, The tokens that are augmented using our TokenAdapt module exhibit more reasonable results, mitigating the image degradation caused by the direct application of pixel-based augmentations to tokens. Consequently, TokenAdapt consistently improves model performance across various scenarios by effectively leveraging pixel-based data augmentation in a token domain. Notably, as shown in Table~\textcolor{red}{2}, we observe that \methodname shows more significant performance gains with MTM, highlighting the importance of minimizing undesired artifacts to overcome the challenges of data augmentation in the tasks where pixel-level information is important.

\subsubsection{ColorAdapt.}
\begin{figure*}[]
    \centering
    \small
    \includegraphics[width=\linewidth]{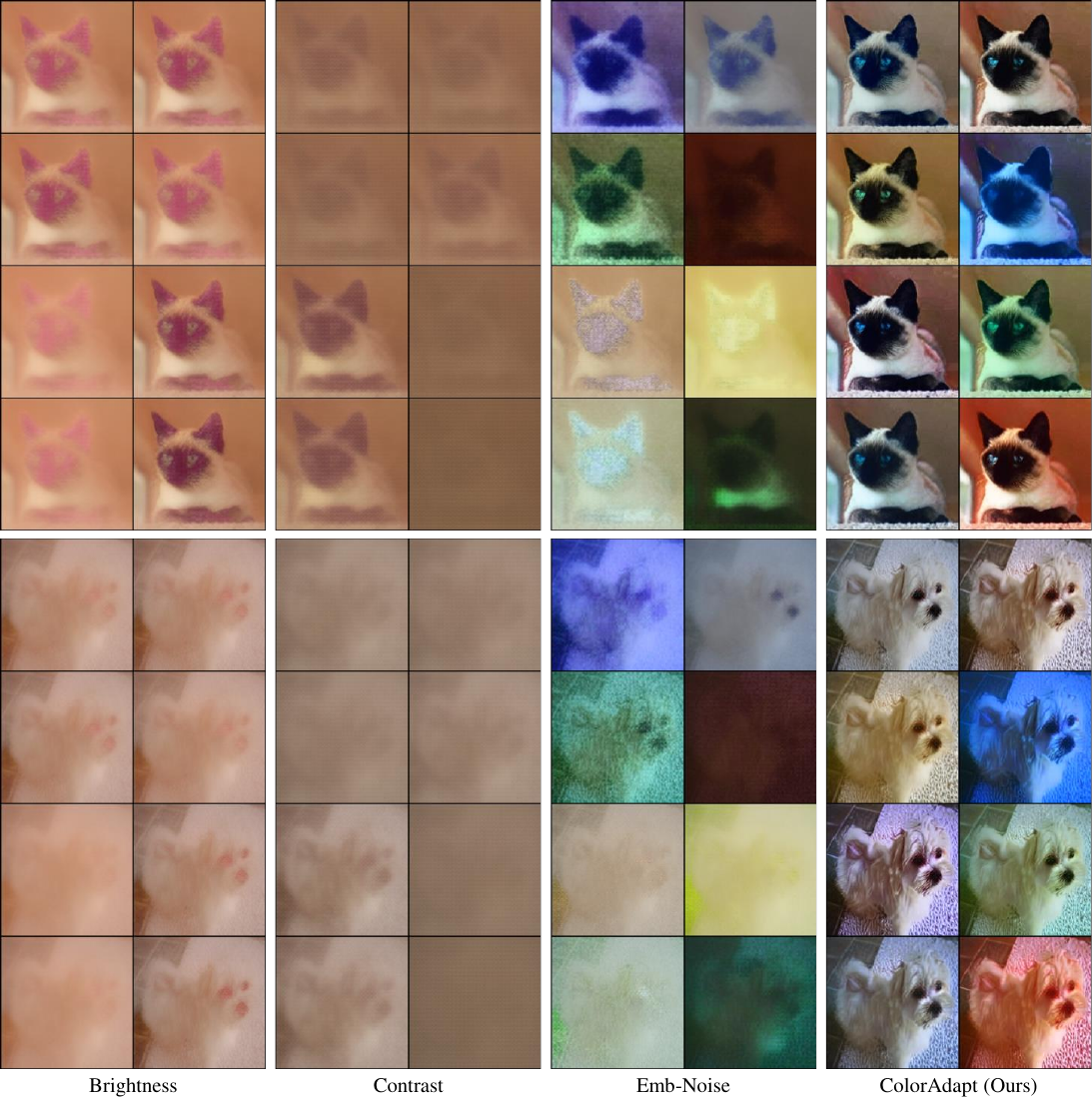}
    \caption{\small \textbf{ColorAdapt provides more reasonable results related to color changes.} We present ViT-VQGAN decoded images to verify the quality of tokenizations after color changes. We use the brightness and contrast function following the implementation~\cite{kornia}. Emb-Noise is the color-based token augmentation~\cite{seit}; we use the same optimized hyper-parameters. Our ColorAdapt effectively preserves object structure in contrast to the failure of the counterparts.}
    \vspace{6em}
    \label{supp:colorAdapt}
\end{figure*}

In Figure~\ref{supp:colorAdapt}, we present the images decoded from color-augmented tokens by various existing color-based data augmentations. We employ brightness and contrast for pixel-based color augmentation, which are widely adopted for pixel-based vision model training. Specifically, we used brightness and contrast functions following the implementation~\cite{kornia}. For token-based color augmentation, we employ Emb-Noise~\cite{seit} with the same optimized hyperparameters used in token-based vision model training~\cite{seit}. Figure~\ref{supp:colorAdapt} illustrates the impact of the proposed ColorAdapt on the visual characteristics of augmented tokens. Unlike the existing color-based augmentations, our ColorAdapt effectively preserves object structure while introducing significant color variations. Recognizing the importance of maintaining object structure in diverse vision tasks (\eg, fine-grained classification, semantic segmentation), our ColorAdapt opens up new possibilities for training more robust and adaptable vision models.

\section{Implementation Details}
\label{appendix:details}

\subsubsection{Masked Token Modeling.}
For masked token modeling (MTM), we follow the training recipe from MAGE~\cite{MAGE}, adjusting the pre-training epochs and the masking ratio. We pre-train the ViT-B model~\cite{ViT} for 400 epochs with a batch size of 4096 and fine-tune the model for 100 epochs with a batch size of 1024. During pre-training, we randomly mask out certain input tokens with a variable masking ratio ranging from 0.4 to 1. The base learning rate is 0.00015 and 0.001 for pre-training and fine-tuning, respectively. In MTM, we replace the Conv $4{\times}4$ Stem-Adapter module from SeiT~\cite{seit} with Conv $2{\times}2$ as the patch embedding layer for ViT models. This adjustment is made because Conv $4{\times}4$ creates overlapping input patches, which hinders representation learning based on masked token modeling. Regarding data augmentation, recognizing the crucial role of pixel-level information in MTM, we apply geometric pixel-data augmentations using only the proposed TokenAdapt module, enhancing our training paradigm's effectiveness.

\subsubsection{Token-based Image Classification.}
For token ImageNet-1k training, we follow the training recipe from SeiT~\cite{seit}, adjusting only the warm-up epochs for stable convergence. We conducted token-based image classification on the ViT-B/16 model~\cite{ViT,deit} and used a learning rate of 0.0015 with cosine scheduling and a weight decay of 0.1. The model was trained for 300 epochs with a batch size of 1024. Regarding data augmentation, SeiT~\cite{seit} incorporates Token-RRC, Token-EDA, Token-CutMix, and Emb-Noise. As a default, we integrate these token-based data augmentation methods and further enhance token diversity by applying additional augmentations using our TokenAdapt and ColorAdapt. Specifically, we apply pixel-based data augmentations (\eg, RRC, hFlip, affine transformations, Mixup~\cite{mixup}, and CutMix~\cite{cutmix}) to tokens using our TokenAdapt module with a probability of 0.5. We adopt the hyperparameters for data augmentation proposed in DeiT~\cite{deit}. For token-based fine-grained classification, following SeiT, we use DeiT's training recipe. When the number of data points decreased in all experiments, we adjust the number of total training iterations to ensure a fair comparison.

\subsubsection{Token-based Semantic Segmentation.}
For the tokenized ADE-20k dataset preparation, we initially resize the entire ADE-20k dataset to $512 {\times} 512$. Following the procedure described in SeiT, we use the ImageNet-1k-trained ViT-VQGAN tokenizer~\cite{ViT-VQGAN} to extract tokens from the resized images. Token-based semantic segmentation on ADE-20k follows the training recipe of mmsegmentation~\cite{mmseg}, using the DeiT-B/16 model with UperNet~\cite{uperNet}. The training involves a learning rate of 6e-5 with polynomial scheduling and a weight decay of 0.01. The model was trained for 80k iterations with a batch size of 16. Regarding data augmentation, we do not employ mixup-based augmentations in token-based semantic segmentation. This behavior is consistent with recent studies~\cite{segformer,segmenter,vit-adapter} that exclude mixup-based augmentation in their training recipes. Furthermore, similar to MTM, we observe that token-level Random Resized Crop (RRC) does not improve performance due to the importance of maintaining structural integrity in pixel-level classification tasks. Thus, we apply geometric pixel-based data augmentations only using our TokenAdapt module.

\end{document}

%% file: table/main_table.tex
\begin{table}[t]
\small
\centering
\caption{\small \textbf{Storage-efficient ImageNet-1k classification.} We report top-1 accuracies (ViT-B) and compression ratio on ImageNet-1k using different data storage reduction methods. Notably, our approach demonstrates substantial storage efficiency compared to competitors.}
\vspace{-0.5em}
\resizebox{.95\linewidth}{!}{
\begin{tabularx}{\textwidth}{Xcrcl}
\toprule
\multicolumn{1}{c}{Method}  & Input                & \# of images & \multicolumn{1}{c}{Top 1 Acc.} & \begin{tabular}[c]{@{}c@{}} Dataset\\ storage size\end{tabular} \\
\midrule
\multirow{1}{*}{Full pixels}         & \multirow{1}{*}{Image}              & 1,281 k   & 81.8      & 140.0 GB   \\ \midrule
\multirow{1}{*}{Uniform random sampling}    & \multirow{4}{*}{Image}             & 512 k   & 74.0       & \hspace{.3em} 54.6 GB \textcolor{darkergreen}{\textbf{(39\%)}}   \\ 
\multirow{1}{*}{C-score~\cite{jiang2020characterizing} based sampling}    &                & 512 k   & 73.3      & \hspace{.3em} 53.3 GB \textcolor{darkergreen}{\textbf{(38\%)}}   \\ 
\multirow{1}{*}{Adjusting JPEG quality factor to 5}    &    & 1,281 k   & 74.6      & \hspace{.3em} 11.0 GB \textcolor{darkergreen}{\textbf{(8\%)}}    \\
\multirow{1}{*}{Adjusting image resolution ($\times$ 0.2)}    &             & 1,281 k   & 75.2 & \hspace{.8em} 9.6 GB \textcolor{darkergreen}{\textbf{(7\%)}}    \\ \midrule
\multirow{1}{*}{SeiT~\cite{seit}}         & \multirow{2}{*}{Token}      & 1,281 k    & 74.0        & \hspace{.4em} \textbf{1.4 GB \textcolor{darkergreen}{\textbf{(1\%)}}} \\
\multirow{1}{*}{\methodname}         &      & 1,281 k    & \textbf{77.8}        & \hspace{.4em} \textbf{1.4 GB \textcolor{darkergreen}{\textbf{(1\%)}}} \\ \bottomrule
\end{tabularx}
\vspace{-2em}
}
\label{tab:main_results}
\end{table}

%% file: table/token_cls.tex
\begin{table}[t]
\small
\centering
\caption{\small \textbf{Token-based ImageNet-1k Classification.} We report top-1 accuracies (ViT-B) on ImageNet-1k. Note that \methodname w/o MTM means the original SeiT~\cite{seit} training with our token augmentation strategies.}
\label{tab:token-cls}
\vspace{-0.5em}
\resizebox{.95\linewidth}{!}{
\begin{tabularx}{0.9\textwidth}{X *{4}{>{\centering\arraybackslash}X}}
\toprule
\multirow{2}{*}{Storage} & \multicolumn{2}{c}{SeiT} & \multicolumn{2}{c}{\methodname} \\ 
\cmidrule(lr){2-3} \cmidrule(lr){4-5}
 & w/o MTM & w/ MTM & w/o MTM & w/ MTM \\
\midrule
1.4 GB & 74.0 & 75.1 & 75.5 \textcolor{darkergreen}{(+1.5)} & \textbf{77.8} \textcolor{darkergreen}{\textbf{(+2.7)}} \\
1.1 GB & 70.2 & 74.1 & 73.1 \textcolor{darkergreen}{(+2.9)} & \textbf{76.5} \textcolor{darkergreen}{\textbf{(+2.4)}} \\
0.8 GB & 66.3 & 70.6 & 69.1 \textcolor{darkergreen}{(+2.8)} & \textbf{74.1} \textcolor{darkergreen}{\textbf{(+3.5)}} \\
0.5 GB & 59.7 & 64.5 & 63.8 \textcolor{darkergreen}{(+4.1)} & \textbf{70.0} \textcolor{darkergreen}{\textbf{(+5.5)}} \\
0.3 GB & 47.2 & 53.9 & 51.2 \textcolor{darkergreen}{(+4.0)} & \textbf{60.6} \textcolor{darkergreen}{\textbf{(+6.7)}} \\
\bottomrule
\end{tabularx}
}
\vspace{-0.5em}
\end{table}

%% file: table/fg_cls.tex
\begin{table}[t]
\small
\centering
\caption{\small \textbf{Fine-grained classification.} We report top-1 accuracies (ViT-B) on various fine-grained datasets. Our method consistently outperforms its counterpart through effective augmentations for tokens.}
\tabcolsep=1em
\label{tab:otherdatasets}
\vspace{-0.5em}
\begin{tabularx}{0.8\textwidth}{X *{4}{>{\arraybackslash}X}}
\toprule
\multicolumn{1}{c}{Method} & \multicolumn{1}{c}{Flowers} & \multicolumn{1}{c}{Cars} & \multicolumn{1}{c}{iNat18} & \multicolumn{1}{c}{iNat19} \\
\midrule
SeiT~\cite{seit}        & 93.5       & 79.7         & 43.1   & 50.1   \\
\methodname & \textbf{95.2 \textcolor{darkergreen}{(+1.7)}}       & \textbf{86.9 \textcolor{darkergreen}{(+7.2)}}         & \textbf{52.4 \textcolor{darkergreen}{(+9.3)}}   & \textbf{58.9 \textcolor{darkergreen}{(+8.8)}}  \\
\bottomrule
\end{tabularx}
\vspace{-1em}
\end{table}

%% file: table/token_seg.tex
\begin{table}[t]
\small
\centering
\caption{\small \textbf{ADE-20k Semantic Segmentation.} We present mIoU results on the ADE-20k validation set, highlighting the effectiveness of our approach in pixel-level classification tasks.}
\label{tab:seg}
\vspace{-0.5em}
\begin{tabularx}{0.75\textwidth}{X *{3}{>{\centering\arraybackslash}X}}
\toprule
\multicolumn{1}{c}{Method} & Backbone & SeiT & \methodname \\
\midrule
UPerNet~\cite{uperNet} & ViT-B~\cite{ViT} & 39.0 & \textbf{43.2 \textcolor{darkergreen}{(+4.2)}} \\
\bottomrule
\end{tabularx}
\end{table}

%% file: table/robustness.tex
\begin{table}[t]
\small
    \centering
    \caption{\small \textbf{Robustness evaluation.} We report top-1 robust accuracies (ViT-B) against corruptions and domain shifts for each model trained on ImageNet-1k. Our method consistently outperforms its counterpart on various robustness benchmarks, indicating that our augmentation strategies enhance the generalizability of trained models.}
    \label{tab:robustness}
\vspace{-1em}
    \resizebox{.95\linewidth}{!}{
    \begin{tabularx}{\textwidth}{X *{6}{>{\arraybackslash}X}}
    \toprule
    \multicolumn{1}{c}{Method} & \multicolumn{1}{c}{Clean} & \multicolumn{1}{c}{Gauss. Noise} & \multicolumn{1}{c}{Gauss. Blur} & \multicolumn{1}{c}{ImageNet-R} & \multicolumn{1}{c}{Sketch} & \multicolumn{1}{c}{ObjectNet} \\ \midrule
    SeiT~\cite{seit}  & 74.0    & 50.7 & 62.6 & 25.5 & 22.6 & 15.7 \\
    \methodname & \textbf{75.5 \textcolor{darkergreen}{\textbf{(+1.5)}}} & \textbf{58.6} \textcolor{darkergreen}{\textbf{(+7.9)}} & \textbf{66.8} \textcolor{darkergreen}{\textbf{(+4.2)}} & \textbf{30.2} \textcolor{darkergreen}{\textbf{(+4.7)}} & \textbf{27.7} \textcolor{darkergreen}{\textbf{(+5.1)}} & \textbf{18.3} \textcolor{darkergreen}{\textbf{(+2.6)}} \\ 
    \bottomrule
    \end{tabularx}
    }
\vspace{-1em}
\end{table}

%% file: table/sub-tables.tex
\begin{table}[t]
\small
\centering
\caption{\small \textbf{Impact of our method on the data-hungry scenario.} We report top-1 accuracies (ViT-S) on ImageNet-100 with varying amounts of training data. Our method benefits more under less training data compared to its counterpart.}
\label{tab:limited_data}

\vspace{-0.5em}
\begin{tabularx}{0.8\textwidth}{X *{4}{>{\arraybackslash}X}}
\toprule
\multicolumn{1}{c}{\# of images} & \multicolumn{1}{c}{127k} & \multicolumn{1}{c}{76k} & \multicolumn{1}{c}{25k} & \multicolumn{1}{c}{13k} \\
\multicolumn{1}{c}{Storage} & \multicolumn{1}{c}{138 MB} & \multicolumn{1}{c}{83 MB} & \multicolumn{1}{c}{28 MB} & \multicolumn{1}{c}{14 MB} \\
\midrule
SeiT~\cite{seit} & 77.3 & 70.3 & 53.3 & 43.2 \\
\methodname & \textbf{81.4 \textcolor{darkergreen}{\textbf{(+4.1)}}} & \textbf{76.5 \textcolor{darkergreen}{\textbf{(+6.2)}}} & \textbf{61.6 \textcolor{darkergreen}{\textbf{(+8.3)}}} & \textbf{52.3 \textcolor{darkergreen}{\textbf{(+9.1)}}} \\
\bottomrule
\end{tabularx}
\end{table}

\begin{table}[t]
\centering
\small
\caption{\small {\bf Impact of the proposed augmentation strategies.} We report the top-1 accuracies (ViT-S) on the ImageNet-100 validation set for various combinations of our token augmentation strategies. Notably, our TokenAdapt and ColorAdapt consistently improve model performance, exhibiting synergy when used together.}
\label{tab:abl_component}
\vspace{-0.5em}
\begin{tabular}{cccc}
\toprule
Framework & ColorAdapt                & TokenAdapt          & Top 1 Acc. \\ \midrule
\multirow{4}{*}{SeiT~\cite{seit}} & \nomark   & \nomark     & 77.3  \\
& \yesmark  & \nomark     & 78.3  \\
& \nomark   & \yesmark    & 80.4  \\
& \yesmark  & \yesmark    & \textbf{81.4}  \\ \bottomrule
\end{tabular}
\end{table}

%% file: table/other_input.tex
\begin{table}[t]
\small
\centering
\caption{\small \textbf{Adaptibility of our method to different input.} Input denotes tokenizers for our model. We report the top-1 accuracies (ViT-S) on ImageNet-100. For VQGAN, we used a publicly available VQGAN~\cite{VQGAN} trained on OpenImages~\cite{openimages} as a tokenizer.}
\label{tab:dct}
\vspace{-0.5em}
\begin{tabular}{ccc}
\toprule
Input                 & SeiT          & \methodname \\ \midrule
\multicolumn{1}{l}{Token (ViT-VQGAN~\cite{ViT-VQGAN})}       & 77.3              & \textbf{81.4 \textcolor{darkergreen}{(+4.1)}}   \\
\multicolumn{1}{l}{Token (VQGAN~\cite{VQGAN})}       & 81.8              & \textbf{83.9 \textcolor{darkergreen}{(+2.1)}}   \\
\multicolumn{1}{l}{DCT coefficients}                              & 69.5              & \textbf{71.0 \textcolor{darkergreen}{(+1.5)}}   \\ \bottomrule
\end{tabular}
\vspace{-1em}
\end{table}

%% file: main.bbl
\begin{thebibliography}{10}
\providecommand{\url}[1]{\texttt{#1}}
\providecommand{\urlprefix}{URL }
\providecommand{\doi}[1]{https://doi.org/#1}

\bibitem{bahng2020learning}
Bahng, H., Chun, S., Yun, S., Choo, J., Oh, S.J.: Learning de-biased representations with biased representations. In: International Conference on Machine Learning. pp. 528--539. PMLR (2020)

\bibitem{Beit}
Bao, H., Dong, L., Piao, S., Wei, F.: Beit: Bert pre-training of image transformers. arXiv preprint arXiv:2106.08254  (2021)

\bibitem{objectnet}
Barbu, A., Mayo, D., Alverio, J., Luo, W., Wang, C., Gutfreund, D., Tenenbaum, J., Katz, B.: Objectnet: A large-scale bias-controlled dataset for pushing the limits of object recognition models. Advances in neural information processing systems  \textbf{32} (2019)

\bibitem{muse}
Chang, H., Zhang, H., Barber, J., Maschinot, A., Lezama, J., Jiang, L., Yang, M.H., Murphy, K., Freeman, W.T., Rubinstein, M., et~al.: Muse: Text-to-image generation via masked generative transformers. arXiv preprint arXiv:2301.00704  (2023)

\bibitem{maskgit}
Chang, H., Zhang, H., Jiang, L., Liu, C., Freeman, W.T.: Maskgit: Masked generative image transformer. In: Proceedings of the IEEE/CVF Conference on Computer Vision and Pattern Recognition. pp. 11315--11325 (2022)

\bibitem{vit-adapter}
Chen, Z., Duan, Y., Wang, W., He, J., Lu, T., Dai, J., Qiao, Y.: Vision transformer adapter for dense predictions. arXiv preprint arXiv:2205.08534  (2022)

\bibitem{chun2020empirical}
Chun, S., Oh, S.J., Yun, S., Han, D., Choe, J., Yoo, Y.: An empirical evaluation on robustness and uncertainty of regularization methods. arXiv preprint arXiv:2003.03879  (2020)

\bibitem{coleman2019selection}
Coleman, C., Yeh, C., Mussmann, S., Mirzasoleiman, B., Bailis, P., Liang, P., Leskovec, J., Zaharia, M.: Selection via proxy: Efficient data selection for deep learning. arXiv preprint arXiv:1906.11829  (2019)

\bibitem{mmseg}
Contributors, M.: {MMSegmentation}: Openmmlab semantic segmentation toolbox and benchmark. \url{https://github.com/open-mmlab/mmsegmentation} (2020)

\bibitem{croce2020reliable}
Croce, F., Hein, M.: Reliable evaluation of adversarial robustness with an ensemble of diverse parameter-free attacks. In: International conference on machine learning. pp. 2206--2216. PMLR (2020)

\bibitem{autoaug}
Cubuk, E.D., Zoph, B., Mane, D., Vasudevan, V., Le, Q.V.: Autoaugment: Learning augmentation strategies from data. In: Proceedings of the IEEE/CVF conference on computer vision and pattern recognition. pp. 113--123 (2019)

\bibitem{randaug}
Cubuk, E.D., Zoph, B., Shlens, J., Le, Q.V.: Randaugment: Practical automated data augmentation with a reduced search space. In: Proceedings of the IEEE/CVF conference on computer vision and pattern recognition workshops. pp. 702--703 (2020)

\bibitem{cubuk2021tradeoffs}
Cubuk, E.D., Dyer, E.S., Lopes, R.G., Smullin, S.: Tradeoffs in data augmentation: An empirical study  (2021)

\bibitem{IMNET-1k}
Deng, J., Dong, W., Socher, R., Li, L.J., Li, K., Fei-Fei, L.: Imagenet: A large-scale hierarchical image database. In: 2009 IEEE conference on computer vision and pattern recognition. pp. 248--255. Ieee (2009)

\bibitem{devlin2018bert}
Devlin, J., Chang, M.W., Lee, K., Toutanova, K.: Bert: Pre-training of deep bidirectional transformers for language understanding. arXiv preprint arXiv:1810.04805  (2018)

\bibitem{cutout}
DeVries, T., Taylor, G.W.: Improved regularization of convolutional neural networks with cutout. arXiv preprint arXiv:1708.04552  (2017)

\bibitem{ViT}
Dosovitskiy, A., Beyer, L., Kolesnikov, A., Weissenborn, D., Zhai, X., Unterthiner, T., Dehghani, M., Minderer, M., Heigold, G., Gelly, S., et~al.: An image is worth 16x16 words: Transformers for image recognition at scale. arXiv preprint arXiv:2010.11929  (2020)

\bibitem{VQGAN}
Esser, P., Rombach, R., Ommer, B.: Taming transformers for high-resolution image synthesis. In: Proceedings of the IEEE/CVF conference on computer vision and pattern recognition. pp. 12873--12883 (2021)

\bibitem{CIM}
Fang, Y., Dong, L., Bao, H., Wang, X., Wei, F.: Corrupted image modeling for self-supervised visual pre-training. arXiv preprint arXiv:2202.03382  (2022)

\bibitem{ford2019adversarial}
Ford, N., Gilmer, J., Carlini, N., Cubuk, D.: Adversarial examples are a natural consequence of test error in noise. arXiv preprint arXiv:1901.10513  (2019)

\bibitem{geirhos2018imagenet}
Geirhos, R., Rubisch, P., Michaelis, C., Bethge, M., Wichmann, F.A., Brendel, W.: Imagenet-trained cnns are biased towards texture; increasing shape bias improves accuracy and robustness. arXiv preprint arXiv:1811.12231  (2018)

\bibitem{goodfellow2014explaining}
Goodfellow, I.J., Shlens, J., Szegedy, C.: Explaining and harnessing adversarial examples. arXiv preprint arXiv:1412.6572  (2014)

\bibitem{mae}
He, K., Chen, X., Xie, S., Li, Y., Doll{\'a}r, P., Girshick, R.: Masked autoencoders are scalable vision learners. In: Proceedings of the IEEE/CVF conference on computer vision and pattern recognition. pp. 16000--16009 (2022)

\bibitem{imagenet-r}
Hendrycks, D., Basart, S., Mu, N., Kadavath, S., Wang, F., Dorundo, E., Desai, R., Zhu, T., Parajuli, S., Guo, M., Song, D., Steinhardt, J., Gilmer, J.: The many faces of robustness: A critical analysis of out-of-distribution generalization. ICCV  (2021)

\bibitem{imagenet-c}
Hendrycks, D., Dietterich, T.: Benchmarking neural network robustness to common corruptions and perturbations. arXiv preprint arXiv:1903.12261  (2019)

\bibitem{adain}
Huang, X., Belongie, S.: Arbitrary style transfer in real-time with adaptive instance normalization. In: Proceedings of the IEEE international conference on computer vision. pp. 1501--1510 (2017)

\bibitem{jia2021scaling}
Jia, C., Yang, Y., Xia, Y., Chen, Y.T., Parekh, Z., Pham, H., Le, Q., Sung, Y.H., Li, Z., Duerig, T.: Scaling up visual and vision-language representation learning with noisy text supervision. In: International conference on machine learning. pp. 4904--4916. PMLR (2021)

\bibitem{jiang2020characterizing}
Jiang, Z., Zhang, C., Talwar, K., Mozer, M.C.: Characterizing structural regularities of labeled data in overparameterized models. arXiv preprint arXiv:2002.03206  (2020)

\bibitem{cars}
Krause, J., Stark, M., Deng, J., Fei-Fei, L.: 3d object representations for fine-grained categorization. In: Proceedings of the IEEE international conference on computer vision workshops. pp. 554--561 (2013)

\bibitem{openimages}
Kuznetsova, A., Rom, H., Alldrin, N., Uijlings, J., Krasin, I., Pont-Tuset, J., Kamali, S., Popov, S., Malloci, M., Kolesnikov, A., et~al.: The open images dataset v4: Unified image classification, object detection, and visual relationship detection at scale. International Journal of Computer Vision  \textbf{128}(7),  1956--1981 (2020)

\bibitem{lee2022dataset}
Lee, S., Chun, S., Jung, S., Yun, S., Yoon, S.: Dataset condensation with contrastive signals. In: International Conference on Machine Learning. pp. 12352--12364. PMLR (2022)

\bibitem{MAGE}
Li, T., Chang, H., Mishra, S., Zhang, H., Katabi, D., Krishnan, D.: Mage: Masked generative encoder to unify representation learning and image synthesis. In: Proceedings of the IEEE/CVF Conference on Computer Vision and Pattern Recognition. pp. 2142--2152 (2023)

\bibitem{swin}
Liu, Z., Lin, Y., Cao, Y., Hu, H., Wei, Y., Zhang, Z., Lin, S., Guo, B.: Swin transformer: Hierarchical vision transformer using shifted windows. In: Proceedings of the IEEE/CVF international conference on computer vision. pp. 10012--10022 (2021)

\bibitem{lopes2019improving}
Lopes, R.G., Yin, D., Poole, B., Gilmer, J., Cubuk, E.D.: Improving robustness without sacrificing accuracy with patch gaussian augmentation. arXiv preprint arXiv:1906.02611  (2019)

\bibitem{madry2017towards}
Madry, A., Makelov, A., Schmidt, L., Tsipras, D., Vladu, A.: Towards deep learning models resistant to adversarial attacks. arXiv preprint arXiv:1706.06083  (2017)

\bibitem{mahajan2018exploring}
Mahajan, D., Girshick, R., Ramanathan, V., He, K., Paluri, M., Li, Y., Bharambe, A., Van Der~Maaten, L.: Exploring the limits of weakly supervised pretraining. In: Proceedings of the European conference on computer vision (ECCV). pp. 181--196 (2018)

\bibitem{flowers}
Nilsback, M.E., Zisserman, A.: Automated flower classification over a large number of classes. In: 2008 Sixth Indian conference on computer vision, graphics \& image processing. pp. 722--729. IEEE (2008)

\bibitem{seit}
Park, S., Chun, S., Heo, B., Kim, W., Yun, S.: Seit: Storage-efficient vision training with tokens using 1\% of pixel storage. arXiv preprint arXiv:2303.11114  (2023)

\bibitem{paul2021deep}
Paul, M., Ganguli, S., Dziugaite, G.K.: Deep learning on a data diet: Finding important examples early in training. Advances in Neural Information Processing Systems  \textbf{34},  20596--20607 (2021)

\bibitem{kornia}
Riba, E., Mishkin, D., Ponsa, D., Rublee, E., Bradski, G.: Kornia: an open source differentiable computer vision library for pytorch. In: Proceedings of the IEEE/CVF Winter Conference on Applications of Computer Vision. pp. 3674--3683 (2020)

\bibitem{rombach2022latentdiffusion}
Rombach, R., Blattmann, A., Lorenz, D., Esser, P., Ommer, B.: High-resolution image synthesis with latent diffusion models. In: Proceedings of the IEEE/CVF Conference on Computer Vision and Pattern Recognition. pp. 10684--10695 (2022)

\bibitem{rosasco2021distilled}
Rosasco, A., Carta, A., Cossu, A., Lomonaco, V., Bacciu, D.: Distilled replay: Overcoming forgetting through synthetic samples. In: International Workshop on Continual Semi-Supervised Learning. pp. 104--117. Springer (2021)

\bibitem{sangermano2022sample}
Sangermano, M., Carta, A., Cossu, A., Bacciu, D.: Sample condensation in online continual learning. In: 2022 International Joint Conference on Neural Networks (IJCNN). pp. 01--08. IEEE (2022)

\bibitem{laion}
Schuhmann, C., Beaumont, R., Vencu, R., Gordon, C., Wightman, R., Cherti, M., Coombes, T., Katta, A., Mullis, C., Wortsman, M., et~al.: Laion-5b: An open large-scale dataset for training next generation image-text models. Advances in Neural Information Processing Systems  \textbf{35},  25278--25294 (2022)

\bibitem{scimeca2021shortcut}
Scimeca, L., Oh, S.J., Chun, S., Poli, M., Yun, S.: Which shortcut cues will dnns choose? a study from the parameter-space perspective. arXiv preprint arXiv:2110.03095  (2021)

\bibitem{singh2022revisiting}
Singh, M., Gustafson, L., Adcock, A., de~Freitas~Reis, V., Gedik, B., Kosaraju, R.P., Mahajan, D., Girshick, R., Doll{\'a}r, P., Van Der~Maaten, L.: Revisiting weakly supervised pre-training of visual perception models. In: Proceedings of the IEEE/CVF Conference on Computer Vision and Pattern Recognition. pp. 804--814 (2022)

\bibitem{augreg}
Steiner, A., Kolesnikov, A., Zhai, X., Wightman, R., Uszkoreit, J., Beyer, L.: How to train your vit? data, augmentation, and regularization in vision transformers. arXiv preprint arXiv:2106.10270  (2021)

\bibitem{segmenter}
Strudel, R., Garcia, R., Laptev, I., Schmid, C.: Segmenter: Transformer for semantic segmentation. In: Proceedings of the IEEE/CVF international conference on computer vision. pp. 7262--7272 (2021)

\bibitem{taori2020measuring}
Taori, R., Dave, A., Shankar, V., Carlini, N., Recht, B., Schmidt, L.: Measuring robustness to natural distribution shifts in image classification. Advances in Neural Information Processing Systems  \textbf{33},  18583--18599 (2020)

\bibitem{deit}
Touvron, H., Cord, M., Douze, M., Massa, F., Sablayrolles, A., J{\'e}gou, H.: Training data-efficient image transformers \& distillation through attention. In: International conference on machine learning. pp. 10347--10357. PMLR (2021)

\bibitem{iNat18}
Van~Horn, G., Mac~Aodha, O., Song, Y., Cui, Y., Sun, C., Shepard, A., Adam, H., Perona, P., Belongie, S.: The inaturalist challenge 2018 dataset. arXiv preprint arXiv:1707.06642  (2018)

\bibitem{iNat19}
Van~Horn, G., Mac~Aodha, O., Song, Y., Cui, Y., Sun, C., Shepard, A., Adam, H., Perona, P., Belongie, S.: The inaturalist challenge 2019 dataset. arXiv preprint arXiv:1707.06642  (2019)

\bibitem{sketch}
Wang, H., Ge, S., Lipton, Z., Xing, E.P.: Learning robust global representations by penalizing local predictive power. In: Advances in Neural Information Processing Systems. pp. 10506--10518 (2019)

\bibitem{wang2018dataset}
Wang, T., Zhu, J.Y., Torralba, A., Efros, A.A.: Dataset distillation. arXiv preprint arXiv:1811.10959  (2018)

\bibitem{EDA}
Wei, J., Zou, K.: Eda: Easy data augmentation techniques for boosting performance on text classification tasks. arXiv preprint arXiv:1901.11196  (2019)

\bibitem{uperNet}
Xiao, T., Liu, Y., Zhou, B., Jiang, Y., Sun, J.: Unified perceptual parsing for scene understanding. In: Proceedings of the European conference on computer vision (ECCV). pp. 418--434 (2018)

\bibitem{segformer}
Xie, E., Wang, W., Yu, Z., Anandkumar, A., Alvarez, J.M., Luo, P.: Segformer: Simple and efficient design for semantic segmentation with transformers. Advances in Neural Information Processing Systems  \textbf{34},  12077--12090 (2021)

\bibitem{simmim}
Xie, Z., Zhang, Z., Cao, Y., Lin, Y., Bao, J., Yao, Z., Dai, Q., Hu, H.: Simmim: A simple framework for masked image modeling. In: Proceedings of the IEEE/CVF Conference on Computer Vision and Pattern Recognition. pp. 9653--9663 (2022)

\bibitem{ViT-VQGAN}
Yu, J., Li, X., Koh, J.Y., Zhang, H., Pang, R., Qin, J., Ku, A., Xu, Y., Baldridge, J., Wu, Y.: Vector-quantized image modeling with improved vqgan. arXiv preprint arXiv:2110.04627  (2021)

\bibitem{cutmix}
Yun, S., Han, D., Oh, S.J., Chun, S., Choe, J., Yoo, Y.: Cutmix: Regularization strategy to train strong classifiers with localizable features. In: Proceedings of the IEEE/CVF international conference on computer vision. pp. 6023--6032 (2019)

\bibitem{mixup}
Zhang, H., Cisse, M., Dauphin, Y.N., Lopez-Paz, D.: mixup: Beyond empirical risk minimization. arXiv preprint arXiv:1710.09412  (2017)

\bibitem{zhao2021dataset}
Zhao, B., Bilen, H.: Dataset condensation with differentiable siamese augmentation. In: International Conference on Machine Learning. pp. 12674--12685. PMLR (2021)

\bibitem{zhao2020dataset}
Zhao, B., Mopuri, K.R., Bilen, H.: Dataset condensation with gradient matching. arXiv preprint arXiv:2006.05929  (2020)

\bibitem{randerasing}
Zhong, Z., Zheng, L., Kang, G., Li, S., Yang, Y.: Random erasing data augmentation. In: Proceedings of the AAAI conference on artificial intelligence. vol.~34, pp. 13001--13008 (2020)

\bibitem{ADE20K}
Zhou, B., Zhao, H., Puig, X., Xiao, T., Fidler, S., Barriuso, A., Torralba, A.: Semantic understanding of scenes through the ade20k dataset. International Journal of Computer Vision  \textbf{127},  302--321 (2019)

\end{thebibliography}
